\documentclass{article}

\usepackage[final]{corl_2020} 
\usepackage{amsmath}
\usepackage{amssymb}
\usepackage{graphicx}
\usepackage[ruled,linesnumbered]{algorithm2e}
\usepackage{booktabs}
\usepackage{float}
\usepackage{bigstrut}
\usepackage{multirow}
\usepackage{balance}
\usepackage{lineno}
\usepackage{amsopn}
\usepackage{comment}
\usepackage{color}
\usepackage{algorithmic}
\usepackage{tabularx}
\usepackage{mathrsfs}
\usepackage{makecell}
\usepackage{url,subfigure,tabulary}
\usepackage[cmintegrals]{newtxmath}
\usepackage{slashbox}

\newcommand{\ie}{\textit{i.e.}}

\usepackage{array}
\makeatletter
\newcommand{\thickhline}{%
    \noalign {\ifnum 0=`}\fi \hrule height 1pt
    \futurelet \reserved@a \@xhline
}
\newcolumntype{"}{@{\hskip\tabcolsep\vrule width 1pt\hskip\tabcolsep}}
\makeatother

\newcolumntype{L}[1]{>{\raggedright\arraybackslash}p{#1}}
\newcolumntype{C}[1]{>{\centering\arraybackslash}p{#1}}
\newcolumntype{R}[1]{>{\raggedleft\arraybackslash}p{#1}}

\usepackage{pifont}
\newcommand{\cmark}{\ding{51}}%

\title{CoT-AMFlow: Adaptive Modulation Network\\ with Co-Teaching Strategy for Unsupervised\\ Optical Flow Estimation}

\author{
  Hengli Wang \\
  ECE Department \\
  HKUST \\
  \texttt{hwangdf@connect.ust.hk} \\
  \And
  Rui Fan \\
  CSE Department \\
  UC San Diego \\
  \texttt{rui.fan@ieee.org} \\
  \And
  Ming Liu \\
  ECE Department \\
  HKUST \\
  \texttt{eelium@ust.hk} \\
}

\begin{document}
\maketitle

\begin{abstract}
    The interpretation of ego motion and scene change is a fundamental task for mobile robots. Optical flow information can be employed to estimate motion in the surroundings. Recently, unsupervised optical flow estimation has become a research hotspot.  However, unsupervised approaches are often easy to be unreliable on partially occluded or texture-less regions. To deal with this problem, we propose CoT-AMFlow in this paper, an unsupervised optical flow estimation approach. In terms of the network architecture, we develop an adaptive modulation network that employs two novel module types, flow modulation modules (FMMs) and cost volume modulation modules (CMMs), to remove outliers in challenging regions. As for the training paradigm, we adopt a co-teaching strategy, where two networks simultaneously teach each other about challenging regions to further improve accuracy. Experimental results on the MPI Sintel, KITTI Flow and Middlebury Flow benchmarks demonstrate that our CoT-AMFlow outperforms all other state-of-the-art unsupervised approaches, while still running in real time. Our project page is available at \url{https://sites.google.com/view/cot-amflow}.
\end{abstract}


\section{Introduction}
\label{sec.introduction}
Mobile robots typically operate in complex environments that are inherently dynamic \cite{ushani2018feature}. Therefore, it is important for such autonomous systems to be conscious of dynamic objects in their surroundings. Optical flow describes pixel-level correspondence between two ordered images, and can be regarded as a useful representation for dynamic object detection. Therefore, many approaches for mobile robot tasks, such as SLAM \cite{zhang2020flowfusion}, dynamic object detection \cite{wang2020atg} and robot navigation \cite{lee2020aggressive}, incorporate optical flow information to improve their performance.

With the development of deep learning technology, deep neural networks have presented highly compelling results for optical flow estimation \cite{dosovitskiy2015flownet,sun2018pwc,hui2018liteflownet}. These networks typically excel at learning optical flow estimation from large amounts of data along with hand-labeled ground truth. However, this data labeling process can be extremely time-consuming and labor-intensive. Recent unsupervised optical flow estimation approaches have attracted much attention, because their advantage in not requiring ground truth enables them to be easily deployed in real-world applications \cite{ren2017unsupervised,meister2018unflow,wang2018occlusion,liu2019ddflow,liu2019selflow}. However, their performance in challenging regions, such as partially occluded or texture-less regions, is often unsatisfactory \cite{wang2018occlusion,liu2020learning}. The underlying cause of this performance degradation is threefold: 1) The popular coarse-to-fine framework \cite{liu2019selflow,liu2020learning} is often sensitive to noises in the flow initialization from the preceding pyramid level, and the challenging regions can introduce errors in the flow estimations, which in turn propagate to subsequent levels. 2) The commonly used cost volume \cite{wang2018occlusion,liu2019ddflow} for establishing feature correspondence can contain many outliers due to the ambiguous correspondence in challenging regions. However, most existing networks directly send the noisy cost volume to the following flow estimation layers without explicitly alleviating the impact of outliers. 3) Many training strategies have been proposed to improve accuracy in challenging regions for unsupervised optical flow estimation, such as occlusion reasoning \cite{meister2018unflow,wang2018occlusion} and self-supervision \cite{liu2019ddflow,liu2019selflow,liu2020learning}. These strategies generally train a single network to provide prior information. However, the prior information is not accurate enough because a single network can be easily disturbed by outliers if the ground truth is inaccessible. Also, the inaccurate prior information can further lead to significant performance degradation.

\begin{figure*}[t]
    \centering
    \includegraphics[width=0.99\textwidth]{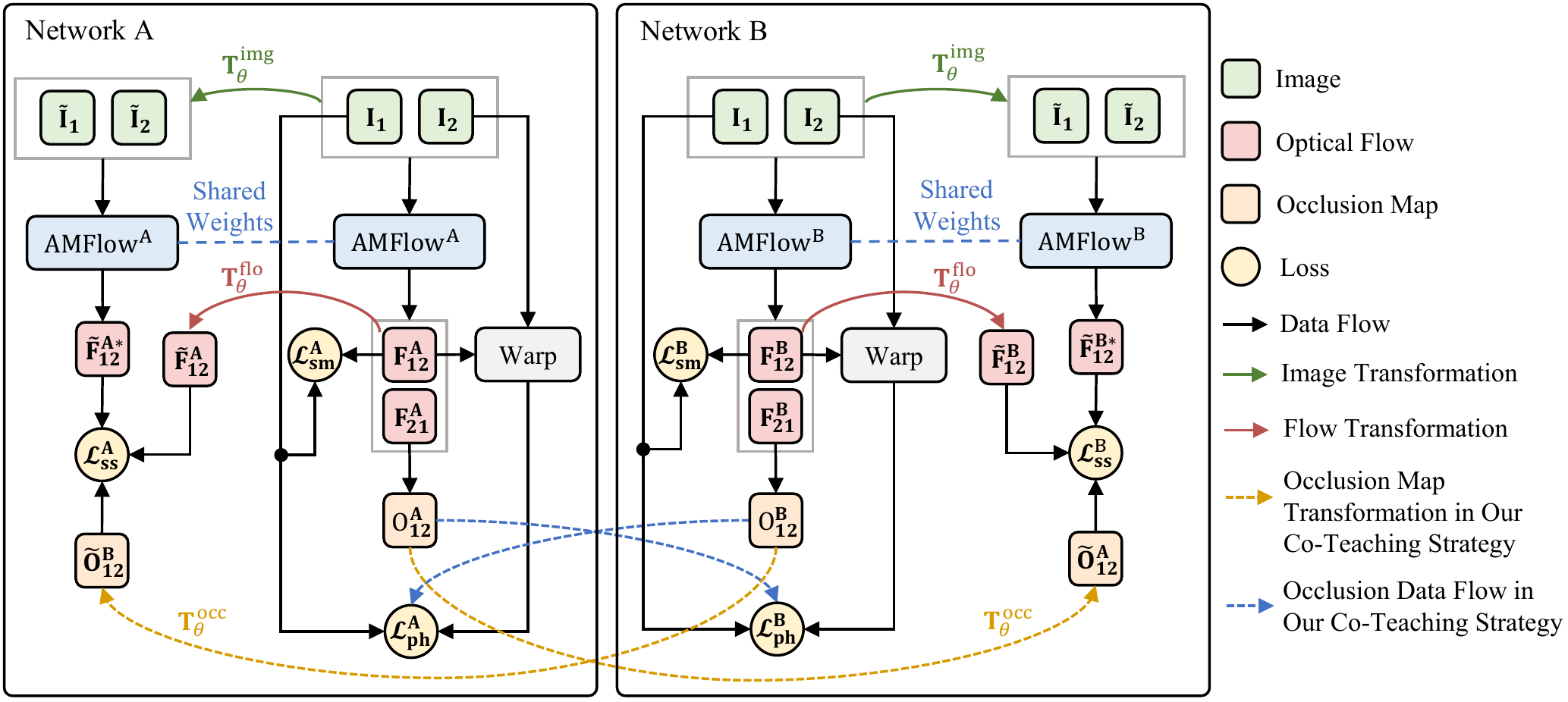}
    \caption{An overview of our CoT-AMFlow. We integrate self-supervision into a co-teaching framework, where two AMFlows with different initializations teach each other about challenging regions to improve stability against outliers and further enhance the accuracy of flow estimation.}
    \label{fig.framework}
\end{figure*}

To overcome these limitations, we propose CoT-AMFlow, which comprises adaptive modulation networks, named AMFlows, that learn optical flow estimation in an unsupervised way with a co-teaching strategy. The overview of our proposed CoT-AMFlow is illustrated in Fig.~\ref{fig.framework}, and we leverage three novel techniques to improve the flow accuracy, as follows:
\begin{itemize}
    \item We apply \textit{flow modulation modules} (FMMs) in our AMFlow to refine the flow initialization from the preceding pyramid level using local flow consistency, which can address the issue of accumulated errors.
    \item We present \textit{cost volume modulation modules} (CMMs) in our AMFlow to explicitly reduce outliers in the cost volume using a flexible and efficient sparse point-based scheme.
    \item We adopt a \textit{co-teaching} strategy, where two AMFlows with different initializations simultaneously teach each other about challenging regions to improve robustness against outliers.
\end{itemize}
We conduct extensive experiments on the MPI Sintel \cite{sintel}, KITTI Flow 2012 \cite{kitti12}, KITTI Flow 2015 \cite{kitti15} and Middlebury Flow \cite{middlebury} benchmarks. Experimental results show that our CoT-AMFlow outperforms all other unsupervised approaches, while still running in real time.

\section{Related Work}
\label{sec.related_work}

\subsection{Optical Flow Estimation}
\label{sec.optical_flow_estimation}
Traditional approaches typically estimate optical flow by minimizing a global energy that measures both brightness consistency and spatial smoothness \cite{horn1981determining,memin1998dense,brox2004high}. With recent development in deep learning technology, supervised approaches using convolutional neural networks (CNNs) have been extensively applied in optical flow estimation, and the achieved results are very promising. FlowNet~\cite{dosovitskiy2015flownet} was the first end-to-end deep neural network for optical flow estimation. It employs a correlation layer to compute feature correspondence. Later on, PWC-Net \cite{sun2018pwc} and LiteFlowNet~\cite{hui2018liteflownet} presented a pyramid architecture, which consists of feature warping layers, cost volumes and flow estimation layers. Such an architecture can achieve remarkable flow accuracy and high efficiency simultaneously. Their subsequent versions \cite{ren2019fusion,hui20liteflownet2} also made incremental improvements. Unsupervised approaches generally adopt similar network architectures to supervised approaches, and focus more on training strategies. However, existing network architectures do not explicitly address the issues of noisy flow initializations and outliers in the cost volume, as previously mentioned. Therefore, we develop the FMMs and CMMs in our AMFlow to overcome these limitations.

Among the training strategies for unsupervised approaches, DSTFlow \cite{ren2017unsupervised} first presented a photometric loss and a smoothness loss for unsupervised training. Additionally, some approaches train a single network to perform occlusion reasoning for accuracy improvement \cite{meister2018unflow,wang2018occlusion}. Self-supervision \cite{liu2019ddflow,liu2019selflow} is also an important strategy for unsupervised training. It first trains a single network to generate flow labels, and then conducts data augmentation to make flow estimations more challenging. The augmented samples are further employed as supervision to train another network. One variant of self-supervision is to train only one network with a two-forward process \cite{liu2020learning}. However, training a single network to provide flow labels is likely to be unreliable due to the disturbance of outliers and the lack of ground-truth supervision. To address this issue, we integrate self-supervision into a co-teaching framework, where two networks simultaneously teach each other about challenging regions to improve stability against outliers.

\subsection{Co-Teaching Strategy}
\label{sec.co-teaching_related}
The co-teaching strategy was first proposed for the image classification task with extremely noisy labels \cite{han2018co}. Since then, many researchers have resorted to this strategy for various specific robust training tasks, such as face recognition \cite{yang2020asymmetric} and object detection \cite{chadwick2019training}. The main difference between previous studies and our approach is that they focus on the task of supervised learning with noisy labels, while we focus on the task of unsupervised learning. Moreover, the noises in their tasks exist at image level (noisy image classification labels), while the outliers in our task exist at pixel level (inaccurate flow estimation pixels in challenging regions).

\section{Methodology}
\label{sec.methodology}

\begin{figure*}[t]
    \centering
    \includegraphics[width=0.99\textwidth]{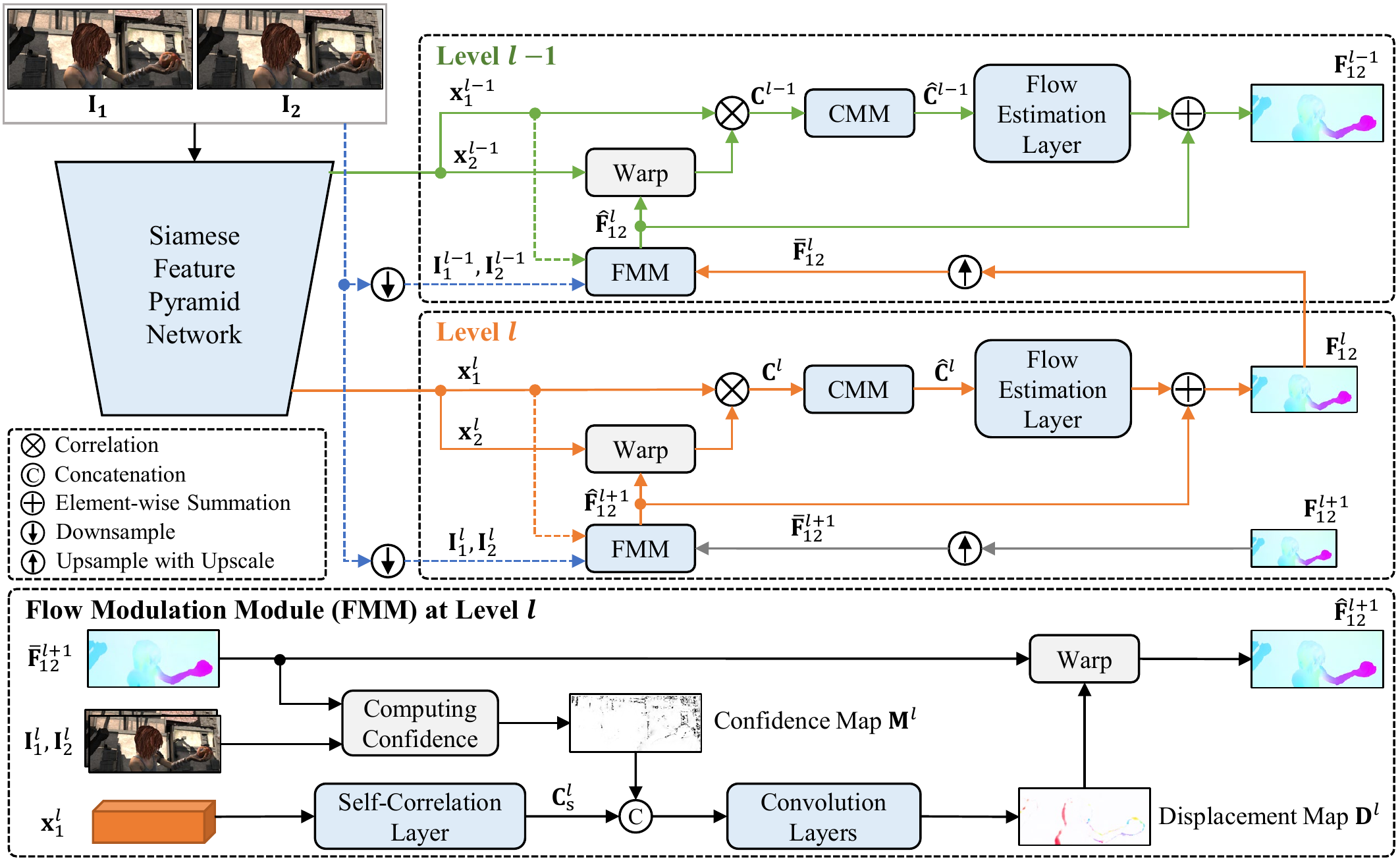}
    \caption{An illustration of our AMFlow, which uses FMMs and CMMs to refine flow initializations and remove outliers in cost volumes, respectively.}
    \label{fig.network}
\end{figure*}

\subsection{AMFlow}
\label{sec.amflow}
In this subsection, we first introduce the overall architecture of our AMFlow, and then present our FMM and CMM. Since we use many notations, we suggest readers refer to the glossary provided in the appendix for better understanding. Fig.~\ref{fig.network} illustrates an overview of our proposed AMFlow, which follows the pipeline of PWC-Net \cite{sun2018pwc}. Different pyramid levels of feature maps are first extracted hierarchically from the input images $\mathbf{I}_{1}$ and $\mathbf{I}_{2}$ using a siamese feature pyramid network, and then are sent to the coarse-to-fine flow decoder. Here, we take level $l$ as an example to introduce our flow decoder, for simplicity. First, the upsampled flow estimation $\overline{\mathbf{F}}_{12}^{l+1}$ at level $l+1$ is processed by our FMM for refinement, and the generated modulated flow $\widehat{\mathbf{F}}_{12}^{l+1}$ is employed to align the feature map $\mathbf{x}_{2}^{l}$ with the feature map $\mathbf{x}_{1}^{l}$. A correlation operation is then employed to compute the cost volume $\mathbf{C}^{l}$, which is then processed by our CMM to remove outliers. After getting the modulated cost volume $\widehat{\mathbf{C}}^{l}$, we take it as input and employ the same flow estimation layer as PWC-Net \cite{sun2018pwc} to estimate the flow residual, which is subsequently added with $\widehat{\mathbf{F}}_{12}^{l+1}$ to obtain the flow estimation $\mathbf{F}_{12}^{l}$ at level $l$. This process iterates and the flow estimations at different scales are generated.

\textbf{Flow Modulation Module (FMM).} In the coarse-to-fine framework, a flow estimation from the preceding level is adopted as a flow initialization at the current level. Therefore, the inaccurate flow estimations in challenging regions can propagate to subsequent levels and cause significant performance degradation. Our FMM is developed to address this problem based on the concept of local flow consistency \cite{zimmer2011optic}.

Our FMM is based on the assumption that the neighboring pixels with similar feature maps should have similar optical flows. Therefore, for a pixel $\mathbf{p}$ with an inaccurate flow estimation $\mathbf{F}\left(\mathbf{p}\right)$, we will look for another pixel $\mathbf{q}$ around $\mathbf{p}$, which has a similar feature map to $\mathbf{p}$ and an accurate flow estimation $\mathbf{F}\left(\mathbf{q}\right)$. Then, we replace $\mathbf{F}\left(\mathbf{p}\right)$ with $\mathbf{F}\left(\mathbf{q}\right)$.

To this end, we first compute a confidence map $\mathbf{M}^{l}$ based on the upsampled flow estimation $\overline{\mathbf{F}}_{12}^{l+1}$ and the downsampled input images $\mathbf{I}_{1}^{l}$ and $\mathbf{I}_{2}^{l}$, as illustrated in Fig.~\ref{fig.network}. The confidence computing operation is defined as follows:
\begin{equation}
    \mathbf{M}^{l} = \exp \left( -\left| \mathcal{B}\left(\mathbf{I}_{1}^{l},\omega\left(\mathbf{I}_{2}^{l}, \overline{\mathbf{F}}_{12}^{l+1} \right) \right) \right| \right),
\end{equation}
where $\mathcal{B}(\cdot,\cdot)$ denotes the function for measuring the photometric difference \cite{liu2020learning}, and $\omega (\mathbf{I},\mathbf{F})$ denotes the warping operation of image $\mathbf{I}$ based on flow $\mathbf{F}$. Then, we use a self-correlation layer to compute a self-cost volume $\mathbf{C}_{\mathbf{s}}^{l}$, which measures the similarity between each pixel in the feature map $\mathbf{x}_{1}^{l}$ and its neighboring pixels. The adopted self-correlation layer is identical to the correlation layer used in the above-mentioned flow decoder, except that it only takes one feature map as input. We further concatenate $\mathbf{M}^{l}$ with $\mathbf{C}_{\mathbf{s}}^{l}$, and send the concatenation to several convolution layers to obtain a displacement map $\mathbf{D}^{l}$. Finally, we warp $\overline{\mathbf{F}}_{12}^{l+1}$ based on $\mathbf{D}^{l}$ to get the modulated flow estimation $\widehat{\mathbf{F}}_{12}^{l+1}$.

\textbf{Cost Volume Modulation Module (CMM).} Ambiguous correspondence in challenging regions can introduce noises into the cost volume, which further influence the subsequent flow estimation layers. Our CMM is designed to reduce noises in the cost volume.

Several traditional approaches have formulated the task of denoising the cost volume as a weighted least squares problem, which obtains the following solution for level $l$ \cite{yoon2006adaptive,hosni2012fast}:
\begin{equation}
    \widehat{\mathbf{C}}^{l}(\mathbf{p}, f)=\sum_{\mathbf{q} \in \mathcal{N}^{l}(\mathbf{p})} w^l(\mathbf{p}, \mathbf{q}) \cdot \mathbf{C}^{l}(\mathbf{q},f),
    \label{eq.traditional_modulation}
\end{equation}
where $\widehat{\mathbf{C}}^{l}(\mathbf{p}, f)$ denotes the modulated cost at pixel $\mathbf{p}$ for flow residual candidate $f$; pixel $\mathbf{q}$ belongs to the neighbors $\mathcal{N}^{l}(\mathbf{p})$ of $\mathbf{p}$; $w^l(\mathbf{p}, \mathbf{q})$ denotes the modulation weight; and $\mathbf{C}^{l}(\mathbf{q},f)$ denotes the original cost at pixel $\mathbf{q}$ for flow residual candidate $f$. Note that the one-dimensional $f$ is transformed from the original two-dimensional flow residual candidate for simplicity, which is the same as the scheme adopted in PWC-Net \cite{sun2018pwc}.

The intuition of our CMM is to implement (\ref{eq.traditional_modulation}) in deep neural networks, which is realized by a flexible and efficient sparse point-based scheme based on deformable convolution \cite{zhu2019deformable}:
\begin{equation}
    \widehat{\mathbf{C}}^{l}(\mathbf{p},f)=\sum_{k=1}^{K} w_{k}^{l} \cdot \mathbf{C}^{l}\left(\mathbf{p}+\mathbf{p}_{k}+\Delta \mathbf{p}_{k}^{l},f\right) \cdot \Delta m_{k}^{l},
    \label{eq.our_modulation}
\end{equation}
where $K$ denotes the number of sampling points; $w_{k}^{l}$ denotes the modulation weight for the $k$-th point; and $\mathbf{p}_{k}$ is the fixed offset of the original convolution layer to $\mathbf{p}$. To make the modulation scheme more flexible, we also employ a separate convolutional layer on $\mathbf{C}^{l}$ to learn an additional offset $\Delta \mathbf{p}_{k}^{l}$ and a spatial-variant weight $\Delta m_{k}^{l}$. These two terms can effectively and efficiently help remove outliers in challenging regions.

\subsection{Loss Function}
\label{sec.loss_function}
\begin{algorithm}[t]
    \KwIn{$\Theta^{\mathrm{A}}$ and $\Theta^{\mathrm{B}}$, learning rate $\eta$, constant threshold $\tau$, epoch $T_k$ and $T_{\mathrm{max}}$, iteration $N_{\mathrm{max}}$.}
    \KwOut{$\Theta^{\mathrm{A}}$ and $\Theta^{\mathrm{B}}$.}
    \For{$T = 1 \to T_{\mathrm{max}}$}
    {
        \textbf{Shuffle} training set $\mathcal{D}$\\
        \For{$N = 1 \to N_{\mathrm{max}}$}
        {
            \textbf{Forward} individually to obtain $\mathbf{F}_{12}^{i}$, $\mathbf{O}_{12}^{i}$, $\tilde{\mathbf{F}}_{12}^{i}$, $\tilde{\mathbf{F}}_{12}^{i*}$ and $\tilde{\mathbf{O}}_{12}^{i}$, $i \in \{\mathrm{A},\mathrm{B}\}$ \\

            \textbf{Set} $\mathbf{O}_{12}^{i} \left( \mathbf{O}_{12}^{i} > \mathcal{R}(T) \right) = 1$, $i \in \{\mathrm{A},\mathrm{B}\}$ \hspace{0.43cm} $\triangleright$ Filter out pixels with high occlusion probability \\

            \textbf{Compute} $\mathcal{L}^{\mathrm{A}} = \mathcal{L}_{\mathrm{ph}}^{\mathrm{A}}(\mathbf{I}_{1}, \mathbf{I}_{2}, \mathbf{F}_{12}^{\mathrm{A}}, \mathbf{O}_{12}^{\mathrm{B}}) + \lambda_{1} \cdot \mathcal{L}_{\mathrm{sm}}^{\mathrm{A}} (\mathbf{I}_{1},\mathbf{F}_{12}^{\mathrm{A}}) + \lambda_{2} \cdot \mathcal{L}_{\mathrm{ss}}^{\mathrm{A}} (\tilde{\mathbf{F}}_{12}^{\mathrm{A}}, \tilde{\mathbf{F}}_{12}^{\mathrm{A}*},\tilde{\mathbf{O}}_{12}^{\mathrm{B}})$\\

            \textbf{Compute} $\mathcal{L}^{\mathrm{B}} = \mathcal{L}_{\mathrm{ph}}^{\mathrm{B}}(\mathbf{I}_{1}, \mathbf{I}_{2}, \mathbf{F}_{12}^{\mathrm{B}}, \mathbf{O}_{12}^{\mathrm{A}}) + \lambda_{1} \cdot \mathcal{L}_{\mathrm{sm}}^{\mathrm{B}} (\mathbf{I}_{1},\mathbf{F}_{12}^{\mathrm{B}}) + \lambda_{2} \cdot \mathcal{L}_{\mathrm{ss}}^{\mathrm{B}} (\tilde{\mathbf{F}}_{12}^{\mathrm{B}}, \tilde{\mathbf{F}}_{12}^{\mathrm{B}*},\tilde{\mathbf{O}}_{12}^{\mathrm{A}})$\\

            \textbf{Update} $\Theta^{i} = \Theta^{i}-\eta \nabla \mathcal{L}^{i}$, $i \in \{\mathrm{A},\mathrm{B}\}$
        }
        \textbf{Update} $\mathcal{R}(T) = 1 - \tau \cdot \min \left\{\frac{T}{T_k}, 1 \right\}$
    }
    \caption{Co-Teaching Strategy}
    \label{alg.co-teaching}
\end{algorithm}

We employ three common loss functions, 1) photometric loss $\mathcal{L}_{\mathrm{ph}}$, 2) smoothness loss $\mathcal{L}_{\mathrm{sm}}$ and 3) self-supervision loss $\mathcal{L}_{\mathrm{ss}}$, to train our CoT-AMFlow, as illustrated in Fig. \ref{fig.framework}. For each network, the forward flow $\mathbf{F}_{12}$ and backward flow $\mathbf{F}_{21}$ can be obtained given the input images $\mathbf{I}_{1}$ and $\mathbf{I}_{2}$. Then, we can compute an occlusion map $\mathbf{O}_{12}$ with the range between 0 and 1 \cite{wang2018occlusion}, where a higher value indicates that the corresponding pixel is more likely to be occluded, and vice versa. Based on these notations, we first introduce our adopted photometric loss \cite{wang2018occlusion} as follows:
\begin{equation}
    \mathcal{L}_{\mathrm{ph}} (\mathbf{I}_{1}, \mathbf{I}_{2}, \mathbf{F}_{12}, \mathbf{O}_{12}) = \frac{\sum_{\mathbf{p}} \psi\left(\mathcal{B}\left( \mathbf{I}_{1},\omega(\mathbf{I}_{2},\mathbf{F}_{12})\right)\right) \odot \left(1-\mathcal{S}\left(\mathbf{O}_{12}\right)\right)}
    {\sum_{\mathbf{p}} \left(1-\mathcal{S}\left(\mathbf{O}_{12}\right)\right)},
\label{eq.ph}
\end{equation}
where $\psi(x)=\sqrt{x^2+0.001^2}$ is the generalized Charbonnier penalty function \cite{sun2010secrets}; $\mathcal{S}(\cdot)$ stands for the stop-gradient; and $\odot$ denotes element-wise multiplication. (\ref{eq.ph}) shows that occluded regions have little impact on $\mathcal{L}_{\mathrm{ph}}$, since there does not exist correspondence in these regions. Moreover, we stop the gradient at the occlusion maps to avoid a trivial solution. Then, the following formulation shows our utilized second-order edge-aware smoothness loss \cite{tomasi1998bilateral}:
\begin{equation}
    \mathcal{L}_{\mathrm{sm}} (\mathbf{I}_{1},\mathbf{F}_{12}) = \frac{1}{N_{\mathbf{p}}} \sum_{\mathbf{p}} \sum_{d \in \{x,y\}} \exp \left(-50 \sum_{c}\left|\frac{\partial \mathbf{I}_{1}}{\partial d}\right|\right) \cdot \left|\frac{\partial^{2} \mathbf{F}_{12}}{\partial d^{2}}\right|,
\label{eq.sm}
\end{equation}
where $c$ denotes the color channel and $N_{\mathbf{p}}$ is the total number of pixels. We also adopt a self-supervision scheme \cite{liu2020learning}. Specifically, we first conduct transformations $\mathbf{T}_{\theta}^{\mathrm{img}}$, $\mathbf{T}_{\theta}^{\mathrm{flo}}$ and $\mathbf{T}_{\theta}^{\mathrm{occ}}$ on $(\mathbf{I}_{1}, \mathbf{I}_{2})$, $\mathbf{F}_{12}$ and $\mathbf{O}_{12}$ respectively to construct augmented samples $\tilde{\mathbf{I}}_{1}$, $\tilde{\mathbf{I}}_{2}$, $\tilde{\mathbf{F}}_{12}$ and $\tilde{\mathbf{O}}_{12}$. The transformations include spatial, occlusion and appearance transformation \cite{liu2020learning}. We also obtain a flow prediction $\tilde{\mathbf{F}}_{12}^{*}$ based on $\tilde{\mathbf{I}}_{1}$ and $\tilde{\mathbf{I}}_{2}$. Then, our self-supervision loss is shown as follows \cite{liu2019ddflow}:
\begin{equation}
    \mathcal{L}_{\mathrm{ss}} (\tilde{\mathbf{F}}_{12}, \tilde{\mathbf{F}}_{12}^{*},\tilde{\mathbf{O}}_{12}) = \frac{\sum_{\mathbf{p}} \psi\left(\left\|\mathcal{S}\left(\tilde{\mathbf{F}}_{12}\right) - \tilde{\mathbf{F}}_{12}^{*} \right\|_{2} \right) \odot \mathcal{S}\left(\tilde{\mathbf{O}}_{12}\right)}
    {\sum_{\mathbf{p}} \mathcal{S}\left(\tilde{\mathbf{O}}_{12}\right)},
\label{eq.ss}
\end{equation}
where $\left\| \cdot \right\|_{2}$ denotes the L2 norm. Note that, different from $\mathbf{O}_{12}$, $\tilde{\mathbf{O}}_{12}$ measures the occlusion relationship between $\tilde{\mathbf{F}}_{12}$ and $\tilde{\mathbf{F}}_{12}^{*}$. A higher value in $\tilde{\mathbf{O}}_{12}$ indicates that the corresponding pixel is less likely to be occluded in $\tilde{\mathbf{F}}_{12}$ but more likely to be occluded in $\tilde{\mathbf{F}}_{12}^{*}$ \cite{liu2019ddflow}. Therefore, (\ref{eq.ss}) shows that $\mathcal{L}_{\mathrm{ss}}$ helps improve the accuracy of flow estimations in challenging regions.

The whole loss function for training our CoT-AMFlow is a weighted sum of the above three losses, as shown on Line 6 and 7 in Algorithm~\ref{alg.co-teaching}. The details will be introduced in Section \ref{sec.co-teaching_our}.

\subsection{Co-Teaching Strategy}
\label{sec.co-teaching_our}
Our co-teaching strategy is illustrated in Fig.~\ref{fig.framework}, and the corresponding steps are shown in Algorithm~\ref{alg.co-teaching}. Specifically, we simultaneously train two networks $\mathrm{A}$ (with parameter $\Theta^{\mathrm{A}}$) and $\mathrm{B}$ (with parameter $\Theta^{\mathrm{B}}$). In each mini-batch, we first let the two networks forward individually to obtain several outputs (Line 4). Then, we filter out the pixels with a high occlusion probability by setting their value in the occlusion map as 1 (completely occluded and thus have no impact on $\mathcal{L}_\mathrm{ph}$) (Line 5). The filtering threshold is controlled by $\mathcal{R}(T)$, which equals 1 at the beginning and then decreases gradually with the increase of epoch number. The key point of our co-teaching strategy is that each network uses the occlusion maps estimated by the other network to compute its own loss function (Line 6 and 7). Finally, we update the parameters of the two networks separately and also update $\mathcal{R}(T)$ (Line~8 and 10). Next, we will answer two important questions about our co-teaching strategy: 1) Why do we need a dynamic threshold $\mathcal{R}(T)$ and 2) why can swapping the occlusion maps estimated by two networks help improve the accuracy for unsupervised optical flow estimation?

To answer the first question, we know that it is meaningless to compute photometric loss on the occluded regions, and thus we adopt an occlusion-masked photometric loss. According to \cite{arpit2017closer}, networks will first learn easy and clear patterns, \ie, unchallenging regions. However, with the number of epochs increasing, networks will gradually be affected by the inaccurately estimated occlusion maps and thus overfit on the occluded regions, which in turn will lead to more inaccurate occlusion estimations and further cause significant performance degradation. To address this, we keep more pixels in the initial epochs, \ie, $\mathcal{R}(T)$ is large. Then, we gradually filter our pixels with high occlusion probability, \ie, $\mathcal{R}(T)$ gradually decreases, to ensure the networks do not memorize these possible outliers.

The dynamic threshold can, however, only alleviate but not entirely avoid the adverse impact of the occluded regions. Therefore, we further adopt a scheme with two networks, which connects to the answer to our second question. The intuition is that different networks have different abilities to learn flow estimation, and correspondingly, they can generate different occlusion estimations. Therefore, swapping the occlusion maps estimated by the two networks can help them adaptively correct the inaccurate occlusion estimations. Compared with most existing approaches that directly transfer errors back to themselves, our co-teaching strategy can effectively avoid the accumulation of errors and thus improve stability against outliers for unsupervised optical flow estimation. Note that since deep neural networks are highly non-convex and a network with different initializations can lead to different local optimums, we employ two AMFlows with different initializations in our CoT-AMFlow, following \cite{han2018co}, as illustrated in Fig.~\ref{fig.framework}.

\section{Experimental Results}
\label{sec.experiment}

\subsection{Dataset and Implementation Details}
\label{sec.dataset}
In our experiments, we set $\lambda_{1}=2$ in our loss function. In addition, we use $\lambda_{2}=0$ for the first 40\% of epochs and increase it to 0.15 linearly for the next 20\% of epochs, after which it stays at a constant value. The learning rate $\eta$ adopts an exponential decay scheme, with the initialization as $10^{-4}$, and the Adam optimizer is used. Moreover, we set $\tau = 0.8$ and $T_{k} = 0.1 T_{max}$ in Algorithm~\ref{alg.co-teaching} for evaluation on public benchmarks.

We first evaluate our CoT-AMFlow on three popular optical flow benchmarks, MPI Sintel \cite{sintel}, KITTI Flow 2012 \cite{kitti12} and KITTI Flow 2015 \cite{kitti15}. The experimental results are shown in Section \ref{sec.benchmark}. Then, we perform a generalization evaluation on the Middlebury Flow benchmark~\cite{middlebury}, as presented in Section \ref{sec.cross}. We also conduct extensive ablation studies to demonstrate the superiority of 1) our selection of $\tau$ and $T_{k}$; 2) our FMM and CMM; 3) our AMFlow over other network architectures; and 4) our co-teaching strategy over other strategies for unsupervised training. The experimental results are presented in the appendix.

Furthermore, we follow a similar training scheme to those of the previous unsupervised approaches \cite{liu2019ddflow,liu2019selflow,liu2020learning} for fair comparison. For the MPI Sintel benchmark, we first train our model on raw movie frames and then fine-tune it on the training set. For the two KITTI Flow benchmarks, we first employ the KITTI raw dataset to pre-train our model and then fine-tune it using multi-view extension data. Additionally, we adopt two standard evaluation metrics, the average end-point error (AEPE) and the percentage of erroneous pixels (F1) \cite{sintel,kitti12,kitti15,middlebury}.

\begin{table*}[t]
    \caption{Evaluation results on the MPI Sintel, KITTI Flow 2012 and KITTI Flow 2015 benchmarks. Here, we show the primary evaluation metrics used on each benchmark. For the Sintel Clean and Final benchmarks, the AEPE (px) for all pixels is presented. For the KITTI Flow 2012 and 2015, ``Noc'' and ``All'' represent the F1 ($\%$) for non-occluded pixels and all pixels, respectively. ``S'' denotes supervised approaches. $^\dagger$ indicates the network using more than two frames. Best results for supervised and unsupervised approaches are both shown in bold font.}
    \centering
    \begin{tabular}{L{3.2cm}C{0.5cm}C{1cm}C{1cm}C{0.8cm}C{0.8cm}C{0.8cm}C{0.8cm}C{1.2cm}}
        \toprule
        \multicolumn{1}{l}{\multirow{2}{*}{Approach}} & \multicolumn{1}{c}{\multirow{2}{*}{S}} & \multicolumn{2}{c}{MPI Sintel} & \multicolumn{2}{c}{KITTI 2012} & \multicolumn{3}{c}{KITTI 2015} \\ \cmidrule(l){3-4} \cmidrule(l){5-6} \cmidrule(l){7-9}
         & & Clean & Final & Noc & All & Noc & All & Time (s) \\ \midrule
        PWC-Net \cite{sun2018pwc} & \cmark & 4.39 & 5.04 & 4.22 & 8.10 & 6.12 & 9.60 & \textbf{0.03} \\
        LiteFlowNet \cite{hui2018liteflownet} & \cmark & 4.54 & 5.38 & 3.27 & 7.27 & 5.49 & 9.38 & 0.09 \\
        LiteFlowNet2 \cite{hui20liteflownet2} & \cmark & 3.48 & 4.69 & 2.63 & 6.16 & 4.42 & 7.62 & 0.05 \\
        MaskFlownet \cite{zhao2020maskflownet} & \cmark & 2.52 & 4.17 & \textbf{2.07} & \textbf{4.82} & 3.92 & 6.11 & 0.06 \\
        RAFT \cite{teed2020raft} & \cmark & \textbf{1.61} & \textbf{2.86} & -- & -- & \textbf{3.07} & \textbf{5.10} & 0.20 \\
        \midrule
        UnFlow \cite{meister2018unflow} & -- & 9.38 & 10.22 & 4.28 & 8.42 & 7.46 & 11.11 & 0.12 \\
        DDFlow \cite{liu2019ddflow} & -- & 6.18 & 7.40 & 4.57 & 8.86 & 9.55 & 14.29 & 0.06 \\
        SelFlow$^{\dagger}$ \cite{liu2019selflow} & -- & 6.56 & 6.57 & 4.31 & 7.68 & 9.65 & 14.19 & 0.09 \\
        ARFlow \cite{liu2020learning} & -- & 4.78 & 5.89 & 4.71 & 8.49 & 8.91 & 11.80 & \textbf{0.01} \\
        ARFlow-mv$^{\dagger}$ \cite{liu2020learning} & -- & 4.49 & 5.67 & 4.56 & \textbf{7.53} & 8.97 & 11.79 & 0.02 \\
        UFlow \cite{jonschkowski2020matters} & -- & 5.21 & 6.50 & 4.26 & 7.91 & 8.41 & 11.13 & 0.04 \\
        \textbf{CoT-AMFlow (Ours)} & -- & \textbf{3.96} & \textbf{5.14} & \textbf{3.50} & 8.26 & \textbf{6.28} & \textbf{10.34} & 0.06 \\ \bottomrule
    \end{tabular}
    \label{tab.benchmark}
\end{table*}

\subsection{Performance on Public Benchmarks}
\label{sec.benchmark}
\begin{figure*}[t]
    \centering
    \includegraphics[width=\textwidth]{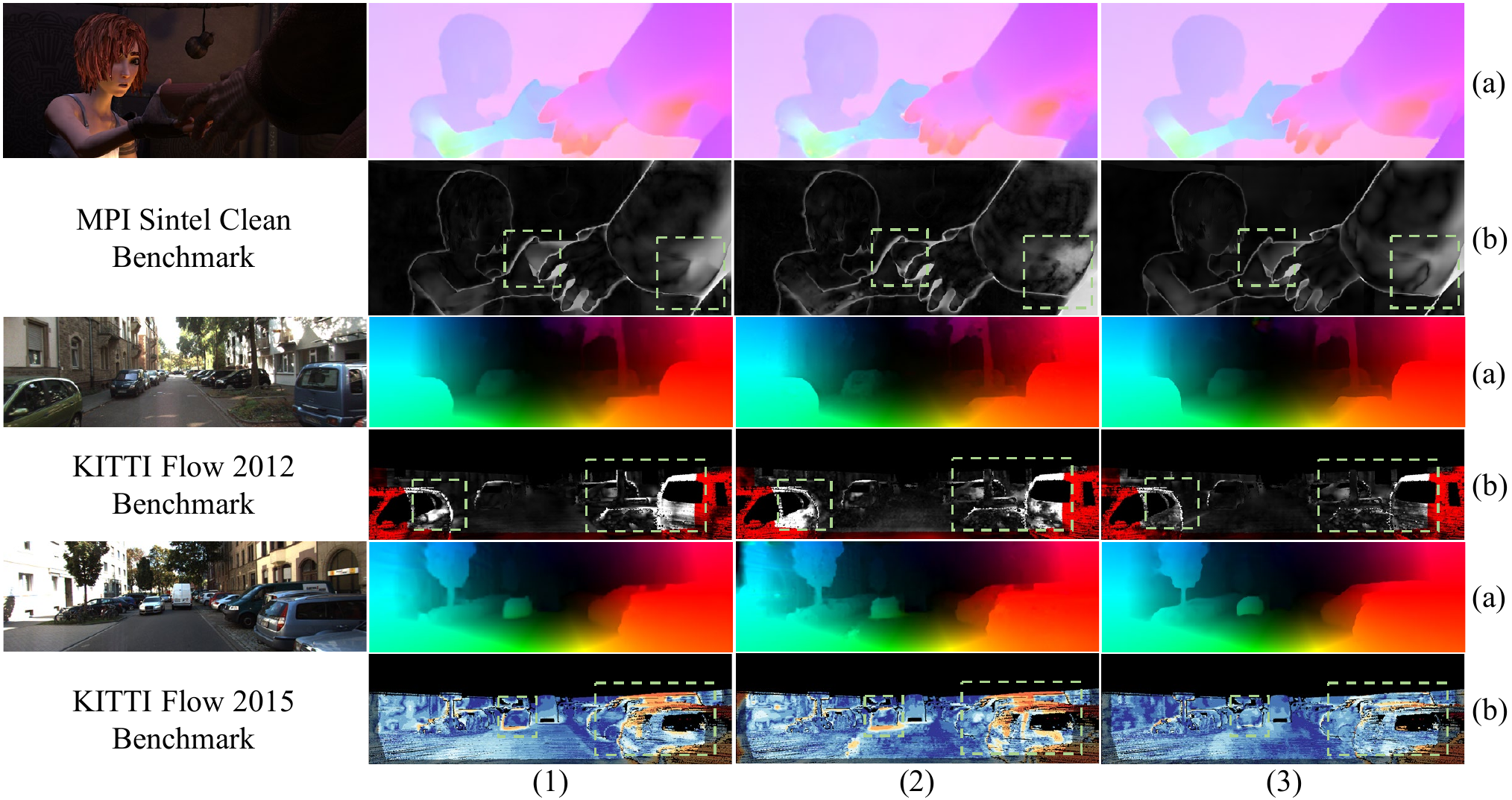}
    \caption{Examples of the MPI Sintel Clean, KITTI Flow 2012 and KITTI Flow 2015 benchmarks, where rows (a) and (b) on columns (1)--(3) show the flow estimations and the corresponding error maps of (1) ARFlow-mv \cite{liu2020learning}, (2) SelFlow \cite{liu2019selflow} and (3) our CoT-AMFlow, respectively. Significantly improved regions are highlighted with green dashed boxes.}
    \label{fig.benchmark}
\end{figure*}

According to the online leaderboards of the MPI Sintel\footnote{\url{http://sintel.is.tue.mpg.de/results}}, KITTI Flow 2012\footnote{\url{http://www.cvlibs.net/datasets/kitti/eval_stereo_flow.php?benchmark=flow}} and KITTI Flow 2015\footnote{\url{http://cvlibs.net/datasets/kitti/eval_scene_flow.php?benchmark=flow}} benchmarks, as shown in Table~\ref{tab.benchmark}, our CoT-AMFlow outperforms all other unsupervised optical flow estimation approaches. We can clearly observe that our approach is significantly ahead of other unsupervised approaches, especially on the MPI Sintel benchmark, where an AEPE improvement of 0.53px--5.42px is achieved on the Sintel Clean benchmark. We also use the KITTI Flow 2015 benchmark to record the average inference time of our CoT-AMFlow. The results in Table~\ref{tab.benchmark} show that our approach can still run in real time with the state-of-the-art performance. One exciting fact is that our unsupervised CoT-AMFlow can achieve considerable performance when compared with supervised approaches. Specifically, on the MPI Sintel Clean benchmark, our CoT-AMFlow outperforms some classic networks such as PWC-Net \cite{sun2018pwc} and LiteFlowNet \cite{hui2018liteflownet}, while achieving only a slightly inferior performance compared with LiteFlowNet2 \cite{hui20liteflownet2}, which demonstrates the effectiveness of our adaptive modulation network and co-teaching strategy. Fig.~\ref{fig.benchmark} illustrates examples of the three public benchmarks, where we can obviously see that our CoT-AMFlow yields more robust and accurate results.

\subsection{Generalization Analysis across Datasets}
\label{sec.cross}
We employ the CoT-AMFlow trained on the MPI Sintel benchmark directly on the Middlebury Flow benchmark to test the generalization ability of our approach. Table~\ref{tab.middlebury} shows the online leaderboard of the Middlebury Flow benchmark\footnote{\url{https://vision.middlebury.edu/flow/eval/results/results-e1.php}}. Note that our CoT-AMFlow has not been fine-tuned on the benchmark. We can observe that our CoT-AMFlow significantly outperforms the unsupervised UnFlow~\cite{meister2018unflow} and even presents superior performance over supervised approaches such as PWC-Net~\cite{sun2018pwc} and LiteFlowNet \cite{hui2018liteflownet}. The results strongly verify that our CoT-AMFlow has an excellent generalization ability.

\begin{table*}[t]
    \caption{Evaluation results on the Middlebury Flow benchmark. ``S'' denotes supervised approaches. Note that our CoT-AMFlow has not been fine-tuned on the benchmark. Best results for supervised and unsupervised approaches are both shown in bold font.}
    \centering
    \begin{tabular}{L{2cm}C{2cm}C{2.7cm}C{1.8cm}C{3.3cm}}
        \toprule
        Metric & PWC-Net \cite{sun2018pwc} & LiteFlowNet \cite{hui2018liteflownet} & UnFlow \cite{meister2018unflow} & \textbf{CoT-AMFlow (Ours)} \\ \midrule
        S & \cmark & \cmark & -- & -- \\
        AEPE (px) & \textbf{0.33} & 0.40 & 0.76 & \textbf{0.26} \\
        \bottomrule
    \end{tabular}
    \label{tab.middlebury}
\end{table*}

\section{Conclusion}
\label{sec.conclusion}
In this paper, we proposed CoT-AMFlow, an adaptive modulation network with a co-teaching strategy for unsupervised optical flow estimation. Our CoT-AMFlow presents three major contributions: 1) a flow modulation module (FMM), which can refine the flow initialization from the preceding pyramid level to address the issue of accumulated errors; 2) a cost volume modulation module (CMM), which can explicitly reduce outliers in the cost volume to improve the accuracy of optical flow estimation; and 3) a co-teaching strategy for unsupervised training, which employs two networks to teach each other about challenging regions to improve robustness against outliers for unsupervised optical flow estimation. Extensive experiments have demonstrated that our CoT-AMFlow achieves the state-of-the-art performance for unsupervised optical flow estimation with an impressive generalization ability, while still running in real time. We believe that our CoT-AMFlow can be directly used in many mobile robot tasks, such as SLAM and robot navigation, to improve their performance. It is also promising to employ the co-teaching strategy in other unsupervised tasks, such as unsupervised disparity or scene flow estimation.

\acknowledgments{We thank the anonymous reviewers for their useful comments. This work was supported by the National Natural Science Foundation of China, under grant No. U1713211, Collaborative Research Fund by Research Grants Council Hong Kong, under Project No. C4063-18G, and HKUST-SJTU Joint Research Collaboration Fund, under project SJTU20EG03, awarded to Prof. Ming Liu.}



\bibliography{egbib}  
\clearpage

{\Large \bf Appendix}

\appendix
This appendix first provides a glossary of notations used in the paper to help readers better follow, as presented in Section \ref{sec.glossary}. Then, we present the experimental results of our ablation studies. Specifically, we first split the original MPI Sintel training set \cite{sintel} into a new training set and a validation set. Then, all models in our ablation studies are trained on the new training set and evaluated on the validation set. In addition, we adopt the average end-point error (AEPE) \cite{sintel} as the evaluation metric.

In our ablation studies, we first explore the impact of different $T_k$ and $\tau$ in the proposed co-teaching strategy on the performance, as presented in Section \ref{sec.tk}. Then, we verify the effectiveness of our flow modulation module (FMM) and cost volume modulation module (CMM) in Section \ref{sec.module}. We also demonstrate the superiority of our AMFlow over other network architectures and the superiority of our co-teaching (CoT) strategy over other strategies for unsupervised training, as presented in Section~\ref{sec.amflow} and Section~\ref{sec.cot}, respectively.

\section{Glossary of Notations}
\label{sec.glossary}
The glossary of notations used in the paper is presented in Table \ref{tab.glossary}.

\section{Impact of Different $T_k$ and $\tau$}
\label{sec.tk}
In our co-teaching strategy, $T_k$ and $\tau$ controls the filtering speed and filtering range of the pixels with high occlusion probability, respectively. We consider three values of $T_k$, $T_k=\{0.05\cdot T_{max},0.10\cdot T_{max},0.15\cdot T_{max}\}$ and five values of $\tau$, $\tau = \{0.70,0.75,0.80,0.85,0.90\}$. We also test the training schemes that adopt a constant $\tau$. The results of our CoT-AMFlow are shown in Table~\ref{tab.cot}. We can observe that the dynamic threshold scheme can effectively improve the performance and our CoT-AMFlow is robust on different choices of $T_k$. Moreover, $\tau$ has a significant impact on the performance. Specifically, a higher $\tau$ indicates that more pixels will be filtered out. We can see that the performance can be improved when $\tau$ increases. However, when too many pixels are filtered out, \ie, $\tau=0.85$ or $\tau=0.90$, the performance can deteriorate because the networks cannot get sufficient training data. Note that we set $T_k = 0.10 \cdot T_{max}$ and $\tau = 0.80$ in the rest of our ablation studies.

\section{Effectiveness of Our FMM and CMM}
\label{sec.module}
Table~\ref{tab.variant} shows the evaluation results of variants of our CoT-AMFlow with some of the proposed modules disabled. We can observe that our FMM and CMM can effectively improve the optical flow accuracy, especially for the pixels with large movements. This is because our FMM can refine the flow initialization from the preceding pyramid level to address the issue of accumulated errors by using local flow consistency, while our CMM can explicitly reduce outliers in the cost volume to improve the accuracy of optical flow estimation by using a flexible and efficient sparse point-based scheme. In addition, the best performance is achieved by integrating our FMM and CMM, which demonstrates the effectiveness of our proposed modules.

\begin{table*}[!t]
    \caption{A glossary of notations used in the paper.}
    \centering
    \begin{tabular}{cl}
    \toprule
       Notation & \multicolumn{1}{c}{Meaning} \\ \midrule
       \multicolumn{2}{c}{Section 3.1} \\ \midrule
       $\mathbf{I}_{1}$ and $\mathbf{I}_{2}$ & The input images \\
       $\mathbf{I}_{1}^{l}$ and $\mathbf{I}_{2}^{l}$ & The downsampled input images at level $l$ \\
       $\mathbf{x}_{1}^{l}$ and $\mathbf{x}_{2}^{l}$ & The feature maps of input images at level $l$ \\
       $\mathbf{F}_{12}^{l}$ & The forward flow estimation at level $l$ \\
       $\overline{\mathbf{F}}_{12}^{l+1}$ & The upsampled forward flow estimation at level $l+1$ \\
       $\widehat{\mathbf{F}}_{12}^{l+1}$ & The modulated forward flow generated via our FMM at level $l+1$ \\
       $\mathbf{M}^{l}$ & The confidence map used in our FMM at level $l$ \\
       $\mathbf{C}_{\mathbf{s}}^{l}$ & The self-cost volume used in our FMM at level $l$ \\
       $\mathbf{D}^{l}$ & The displacement map used in our FMM at level $l$ \\
       $\mathbf{C}^{l}$ & The cost volume at level $l$ \\
       $\widehat{\mathbf{C}}^{l}$ & The modulated cost volume generated via our CMM at level $l$ \\ \midrule
       \multicolumn{2}{c}{Section 3.2 and 3.3} \\ \midrule
       $\mathbf{I}_{1}$ and $\mathbf{I}_{2}$ & The input images\\
       $\mathbf{F}_{12}$ & The forward flow estimation \\
       $\mathbf{F}_{21}$ & The backward flow estimation \\
       $\mathbf{O}_{12}$ & The occlusion map \\
       $\mathbf{T}_{\theta}^{\mathrm{img}}$, $\mathbf{T}_{\theta}^{\mathrm{flo}}$ and $\mathbf{T}_{\theta}^{\mathrm{occ}}$ & The transformations employed on $\mathbf{I}_{1}$, $\mathbf{I}_{2}$, $\mathbf{F}_{12}$ and $\mathbf{O}_{12}$, respectively \cite{liu2020learning} \\
       $\tilde{\mathbf{I}}_{1}$, $\tilde{\mathbf{I}}_{2}$, $\tilde{\mathbf{F}}_{12}$ and $\tilde{\mathbf{O}}_{12}$ & The samples augmented via the above-mentioned transformations \\
       $\tilde{\mathbf{F}}_{12}^{*}$ & The forward flow prediction based on $\tilde{\mathbf{I}}_{1}$ and $\tilde{\mathbf{I}}_{2}$ \\ \bottomrule
    \end{tabular}
    \label{tab.glossary}
\end{table*}

\begin{table*}[!t]
    \caption{AEPE (px) results of our CoT-AMFlow with different $T_k$ and $\tau$ in the proposed co-teaching strategy. The best result is shown in bold font.}
    \centering
    \begin{tabular}{lccccc}
    \toprule
        & $\tau = 0.70$ & $\tau = 0.75$ & $\tau = 0.80$ & $\tau = 0.85$ & $\tau = 0.90$ \\ \midrule
        Constant $\tau$ & 4.31 & 4.10 & 3.95 & 4.34 & 4.89 \\
        $T_k = 0.05 \cdot T_{max}$ & 4.22 & 3.98 & 3.83 & 4.16 & 4.65 \\
        $T_k = 0.10 \cdot T_{max}$ & 4.27 & 4.05 & \textbf{3.79} & 4.02 & 4.64 \\
        $T_k = 0.15 \cdot T_{max}$ & 4.29 & 3.92 & 3.85 & 4.13 & 4.51 \\ \bottomrule
    \end{tabular}
    \label{tab.cot}
\end{table*}

\begin{table*}[!t]
    \caption{AEPE (px) results of variants of our CoT-AMFlow with some of the proposed modules disabled, where ``All'' denotes the AEPE over all pixels, and ``$\mathrm{s}0-10$'', ``$\mathrm{s}10-40$'' and ``$\mathrm{s}40+$'' denote the AEPE over pixels that move less than 10 pixels, between 10 and 40 pixels and more than 40 pixels, respectively. Best results are shown in bold font.}
    \centering
    \begin{tabular}{cccccc}
    \toprule
    FMM & CMM & All & $\mathrm{s}0-10$ & $\mathrm{s}10-40$ & $\mathrm{s}40+$ \\ \midrule
    -- & -- & 4.73 & 0.82 & 2.46 & 29.75 \\
    \cmark & -- & 4.12 & 0.79 & 2.32 & 25.12 \\
    -- & \cmark & 4.23 & \textbf{0.73} & 2.23 & 26.49 \\
    \cmark & \cmark & \textbf{3.79} & 0.76 & \textbf{2.07} & \textbf{23.10} \\\bottomrule
    \end{tabular}
    \label{tab.variant}
\end{table*}

\section{Superiority of Our AMFlow over Other Network Architectures}
\label{sec.amflow}
To further demonstrate the superiority of our AMFlow over other network architectures, we compare the performance of different combinations of unsupervised network architectures and unsupervised training strategies. The results are shown in Table~\ref{tab.comparison}. From rows a)--d), we can observe that for each existing unsupervised approach, the performance can be significantly improved when the network architecture is changed from the original one to our AMFlow, which strongly demonstrates the effectiveness of our architecture. The reason why our AMFlow performs better is that it can address the issues of accumulated errors and reduce outliers in the cost volume to improve the optical flow accuracy by using our FMMs and CMMs. Moreover, from row e), we can see that, compared with other network architectures, our AMFlow achieves the best performance when equipped with the same training strategy, \ie, our co-teaching strategy, which further demonstrates the superiority of our AMFlow over other network architectures.

\section{Superiority of Our Co-Teaching Strategy over Other Strategies for Unsupervised Training}
\label{sec.cot}
From columns 1)--4) in Table~\ref{tab.comparison}, we can observe that for each existing unsupervised approach, the performance can be significantly improved when the training strategy is changed from the original one to our co-teaching strategy, which strongly demonstrates the effectiveness of our strategy. The reason why our co-teaching strategy performs better is that it can improve robustness against outliers for unsupervised optical flow estimation by employing two networks to teach each other about challenging regions simultaneously. Moreover, from column 5), we can see that, compared with other training strategies, our co-teaching strategy achieves the best performance when employed in the same network architecture, \ie, our AMFlow, which further demonstrates the superiority of our co-teaching strategy over other strategies for unsupervised training.

\begin{table*}[!t]
    \caption{AEPE (px) results of different combinations of unsupervised network architectures and unsupervised training strategies. Note that XXXNet and XXXStrat denote the corresponding network architecture and unsupervised training strategy used in XXX, respectively. $^\dagger$ indicates the network using more than two frames. The best result is shown in bold font.}
    \centering
    \begin{tabular}{l|ccccc}
    \noalign{\hrule height 1pt}
    \backslashbox{Strategy}{Network}
     & \makecell{1)UnFlow-\\Net \cite{meister2018unflow}} & \makecell{2)DDFlow-\\Net \cite{liu2019ddflow}} & \makecell{3)SelFlow-\\Net$^\dagger$ \cite{liu2019selflow}} & \makecell{4)ARFlow-\\Net \cite{liu2020learning}} & \makecell{5)\textbf{AMFlow}\\\textbf{(Ours)}} \\ \noalign{\hrule height 0.5pt}
    a) UnFlowStrat \cite{meister2018unflow} & 8.87 & -- & -- & -- & 6.61 \\
    b) DDFlowStrat \cite{liu2019ddflow} & -- & 5.95 & -- & -- & 5.59 \\
    c) SelFlowStrat \cite{liu2019selflow} & -- & -- & 5.22 & -- & 4.98 \\
    d) ARFlowStrat \cite{liu2020learning} & -- & -- & -- & 4.67 & 4.36 \\
    e) \textbf{Co-Teaching (Ours)} & 5.65 & 4.73 & 3.94 & 4.29 & \textbf{3.79} \\\noalign{\hrule height 1pt}
    \end{tabular}
    \label{tab.comparison}
\end{table*}

\end{document}